# Using depth information and colour space variations for improving outdoor robustness for instance segmentation of cabbage

Nils Lüling[1], David Reiser[1], Alexander Stana[1], H.W. Griepentrog[1]

*Abstract*— Image-based yield detection in agriculture could raise harvest efficiency and cultivation performance of farms. Following this goal, this research focuses on improving instance segmentation of field crops under varying environmental conditions. Five data sets of cabbage plants were recorded under varying lighting outdoor conditions. The images were acquired using a commercial mono camera. Additionally, depth information was generated out of the image stream with Structure-from-Motion (SfM). A Mask R-CNN was used to detect and segment the cabbage heads. The influence of depth information and different colour space representations were analysed. The results showed that depth combined with colour information leads to a segmentation accuracy increase of 7.1%. By describing colour information by colour spaces using light and saturation information combined with depth information, additional segmentation improvements of 16.5% could be reached. The CIELAB colour space combined with a depth information layer showed the best results achieving a mean average precision of 75.

## I. Introduction

Nowadays, the main costs for vegetable production are due to labour costs. Especially the harvesting of vegetables is still mainly done by hand. Additionally, it is becoming more challenging to find qualified and affordable workers in developed countries [1]. Therefore, there is great interest in automating this process. For automating the harvest or optimising the logistics, sensor-based detection is essential. Due to a large amount of work involved in vegetable production, working hours into the night are not uncommon. Therefore, robust sensor detection systems are necessary to ensure that the work can be carried out in the future with intelligent machines.

Instance segmentation could offer the possibility of identifying individual plants as well as their fruit by camera images. The analysis could be used to estimate the yield and the best time for harvest. By knowing the plant's position in a two-dimensional image and the additional acquisition of depth information, the plant could be located in a three-dimensional space and harvested by robots [2]. In order to test the performance of instance segmentation, cabbage provides a good case study, as there are several aggravating factors. The green cabbage fruit must be segmented from the green leaves of the plant, as can be seen in Fig. 1 (a) and Fig. 4. The cabbage leaves cover a large part of the fruit, which makes the segmentation challenging.

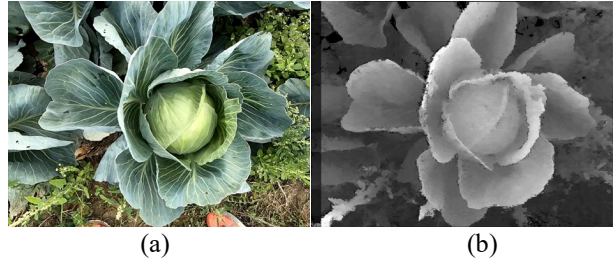

Fig.1: Image comparison. (a) RGB-Image (b) Depth-Image

The environment on outdoor fields is constantly varying. Based on weather, illumination and location, it is difficult to find the same conditions during the data acquisition periods. On the other hand, actively providing shade and illuminating the crop artificially in the field is unpractical and hard to implement. Also the terrain mobility of the sensor carrier platform gets limited. Using depth information could improve the robustness of the segmentation because it is, in general, less dependent on light (Fig.1(b)). Alternative colour representations and colour spaces could also improve the detection algorithms' light independence [3]. This work can serve as a basis for further supporting measures to implement a robust and viable instance segmentation application for field vegetables.

## II. Related Work

There is a great deal of research in the field of automated detection and segmentation of plants [4]–[8]. Most of the work focuses on fruit detection for harvesting or segmentation for weed control [5], [7]. Many research approaches are trying to improve detection and segmentation robustness to achieve consistent performance under the demanding environmental conditions in agriculture caused by light, fog and dust. The possibilities for improving detection and segmentation results can be divided into three sections.

The first option is the choice and number of sensors. Most approaches are based on colour and depth information from a Microsoft Kinect camera [9]. To be more independent of external influences, Milioto *et al.* [7] and Bender *et al.* [8] use artificial shading and exposure, which results in more uniform exposure rates. Bender *et al.* [8] even include near-infrared, colour, 3D, and thermal information in their data set for detection. The detection and segmentation accuracy can be increased by a large number and an intelligent fusion of different sensors. Gai *et al.* [9] show that a fusion of colour and depth information of a Microsoft Kinect camera increases the segmentation accuracy.

[1] All Authors are with the University of Hohenheim, Institute of Agricultural Engineering, Garbenstraße 9, 70599 Stuttgart, Germany. Nils.lueling@uni-hohenheim.de

The second way to improve robustness is to optimise or adapt the segmentation process. Gai et al. [9] improved the watershed instance segmentation of broccoli and iceberg lettuce by manually creating features. An alternative approach to the sensor-based acquisition of depth information would be a neural network-based approach with a multi-task learning concept [11]. Lu *et al.* [12] show that an unsupervised multi-task learning approach can generate depth information and could be used to improve semantic segmentation. Adjustments to the respective application's segmentation procedure are essential in achieving high degrees of precision and high performing speeds.

The third step of making detection and segmentation networks more robust lies in pre-processing the data before it is passed on to an evaluation algorithm. Blok *et al.* [10] reduced the effort of training image creation by data augmentation with geometric and photometric transformations. Data augmentation positively influences segmentation accuracy, as segmentation can be made more independent of perspective, colour intensity, or background [13]. The problem with image augmentation is that the reproduction of images of different exposure does not create new shadows, thus, cannot be robust concerning new shading variations. Another approach to providing information-rich data is to analyse the recorded colour space. In their approach, Diaz-Cely *et al.* [14] stated that different colour spaces achieve different detection results depending on the network architecture. To the authors' knowledge, no work was performed that improves the instance segmentation of cabbage heads under different exposure scenarios due to the influence of depth information as well as colour space variations.

### III. APPROACH

This paper aims to use a Mask R-CNN to accurately detect and segment cabbage heads in images and evaluate the influence of depth information and different colour space representations under varying exposure scenarios. Without relying on expensive and complex sensor technology and an elaborately created training data set representing all exposure scenarios, this approach represents a simple way of improving the robustness of an instance segmentation.

The application case of detecting and analysing field crops by a vehicle driving through the rows leads to high information content in calculating a depth image. The linear movement of the camera over the rows leads to a more reliable rectification of the images. When there is no wind, a static environment is created, making it possible to calculate disparity values at different points in time. Besides the possibility of the camera's high recording speed means that various working speeds can be performed without worrying about too little or too much disparity. The resulting depth images can serve as an additional image layer to the colour information as an input to the neural network.

In order to also affect the exposure influence through the colour information, the effect of different colour spaces on the segmentation accuracy was investigated. By analysing the colour channels of different colour spaces (RGB, CieLab, HSV), providing colour, light, and saturation information on different channels, the neural network can prioritise colour and exposure information differently.

#### A. Experimental Setup

The recordings for training and testing were taken near Stuttgart, Germany, two weeks before the harvest. There was no additional mechanic stabilization, shading, or artificial exposure of the camera. A GoPro camera (Hero 7, GoPro Inc., San Mateo, CA, USA) with a recording speed of 60 frames per second and a resolution of 1920x1440 pixels was used. The camera was moved across the row at a speed of about 1 m/s and a constant height of 90 cm. For the analysis of cabbage, a vertical perspective of the camera to the plant was used. Especially at late growth stages, even a slight tilt of the camera leads to a highly restricted view of the cabbage head due to the cabbage's large vertical leaves. The recordings for training and test images were taken in two different areas in the field. Five series of shots were taken. The first series of images was taken at 09:00 am (L1) in diffuse light (shallow cloud cover), resulting in no shadows in the images. The second series of photographs was taken at 10 am (L2), the third at 1:30 pm with the highest sun (L3), the fourth at 3:30 pm with light cloud cover (L4), and the fifth at 6:30 pm (L5). Due to the low position of the sun, L5 showed the most intense shadows. Fig.4 shows the same cabbage plant at different shooting times. By photographing the same cabbages at different sun elevations, an attempt was made to isolate the influence of exposure. In order to be able to analyse the influence of the different exposure situations more precisely, the instance segmentation network was only trained with diffusely exposed images. An area for training images and an area for test images was defined to avoid overlapping of training and test cabbage images on the field. For the evaluation of the different approaches, 60 images were taken randomly from each test data set. The mean average precision of common objects in context (COCO) detection evaluation matrix was used as the evaluation parameter for the segmentation performance [20]. The mean average precision is calculated via precision (1) and recall (2). For this purpose, the ratio of true positives (TP), false positives (FP), and false negatives (FN) are determined. To calculate the average precision (AP), a TP is evaluated from an intersection over union (IoU) value of 0.5. For the mean average precision (mAP), the AP result is averaged over the IoU steps from 0.5 to 0.95 in 0.05 steps.

$$Precision = \frac{TP}{TP + FP} \quad (1)$$

$$Recall = \frac{TP}{TP + FN} \quad (2)$$

## A. Software Setup

As an instance segmentation network, the Mask R-CNN method with a Resnet50 backbone was used [21]. The Mask R-CNN is based on Tensorflow (GPU 1.3.0) and the Keras library (2.0.8) and is implemented in Python (3.6). The network was trained with 2000 training images of cabbage with a resolution of 1024x1024 pixels. The complete network was trained with randomly weighted start weights and a learning rate of 0.0001 for 10 epochs. No transfer learning approach was used, as existing data sets such as the Microsoft COCO data set are based on RGB images and would have distorted the result compared to other colour spaces. The entire calculation of depth information was done with Matlab (Matlab R2020a, The MathWorks Inc., Natick, MA, USA). The training and the depth image calculation were carried out on a Workstation with an AMD Ryzen Threadripper 2920X 12-core processor (64 GB RAM) and an 8 GB graphics card (GeForce RTX 2070 SUPER).

## B. Mask R-CNN

The Mask R-CNN is an extension of the Faster R-CNN and provides three outputs. A classification is provided for each object to be detected as well as a bounding box. The difference to detection by a Faster R-CNN is the third output. The Mask R-CNN provides an object mask for each detected object, making instance segmentation possible. This mask could then be used for further calculations, e.g., for calculations of the cabbage volume; this segmented mask is crucial, as it can provide information about the cabbage head's cross-section. On the other hand, a simple detection by a Faster R-CNN would allow a more robust implementation but provides too little information for a more precise analysis of the cabbage head. Fig.2 shows the three outputs of the Mask R-CNN. The first step of the Mask R-CNN is the provision of multiple regions of interest (ROI) by a region proposal network (RPN) (Fig.2 (a)). A final detection is provided by procedures such as bounding box refinement and non-maximum suppression. Parallel to the detection, a binary mask of the detection area is created by a fully connected network (FCN) (Fig.2(b) [16]. Fig. 2(c) shows the combination of these three outputs (bounding box, classification and segmentation).

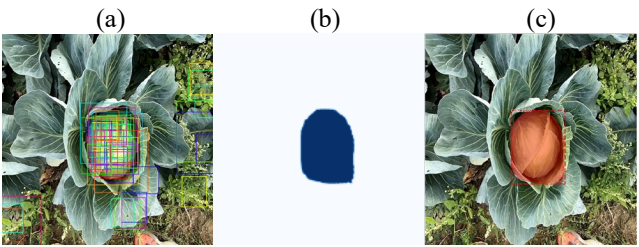

Fig.2: Instance segmentation process: (a) Create bounding boxes, (b) Create binary mask, (c) Combine bounding box, classification and binary mask.

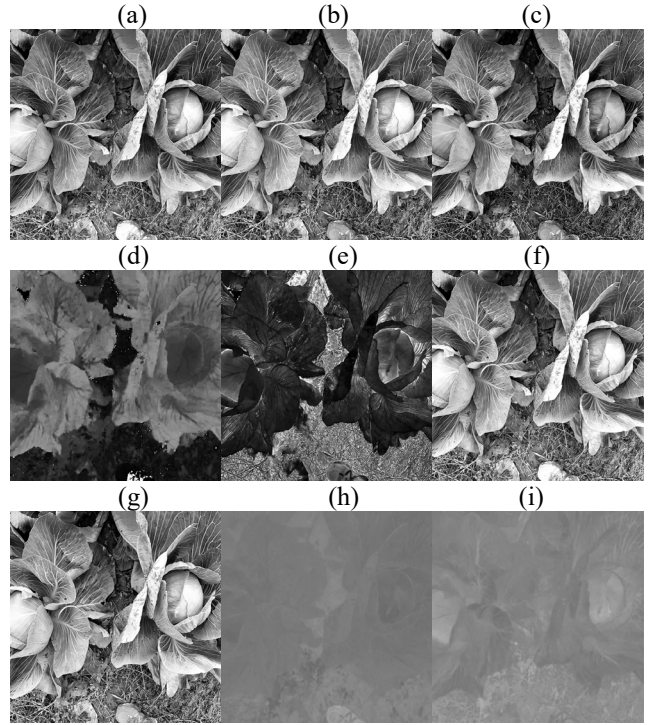

Fig.3: Image channel comparison of RGB (a-c), HSV (d-f) and CIELAB (g-i): (a) Red, (b) Green, (c) Blue, (d) Hue, (e) Saturation, (f) Value (Light), (g) Lightness, (h) A (green to red), (i) B (blue to yellow)

## C. Colour space variations

For analysing the influence of different colour spaces, three colour spaces were used. RGB, CIELAB and HSV. These three colour spaces were chosen for their different colour information provision. The RGB colour space has three colour channels (red, green, and blue). The CIELAB colour space has two colour channels (A and B) and one exposure channel (L). The HSV colour space has only one colour channel (H), one exposure channel (V) and one colour saturation channel (S). Fig. 3 shows the different channels of the colour spaces. The segmentation results can then be used to determine whether relevant information is lost when compressing it to two or one colour channel and whether new information can be gained by providing exposure or saturation.

## D. Depth Image Estimation

The depth information calculation method described here is the same SfM approach used in our other paper on this topic to calculate the volume of cabbage heads as well as the leaf area of single cabbage leaves [23]. A process consisting of four steps was necessary to calculate the depth information.

1. To find anchor points for orientation, speeded-up-robust-features (SURF) [17] were detected and matching features were extracted [18].

2. With the extracted features, the fundamental matrix that is necessary for the rectification could be calculated. The next step was the calculation of the transformation matrices for the image rectification [22].
3. A semi-global matching procedure was used, that generates a disparity image from the two orientated images [19].
4. To further improve the depth image, the disparity of the temporally previous and following image were calculated out of the video stream. As a result, two disparity images are available for the initial image. By merging these two disparity images, a higher depth image quality could be achieved [23].

## IV. RESULTS AND DISCUSSION

To evaluate the intensity of the different exposures and the resulting shadows in the dataset, the standard deviation of the L-value of the CIELAB colour scale was calculated. The results are shown in Table I. The L value indicates the brightness of the pixel from 0 to 100. A high standard deviation of the value stands for a strong shadow cast, as the direct influence of light creates many bright areas and the resulting shadows many dark areas. The highest standard deviation could be seen in dataset L5 with 73.5.

By comparing the results of Table II with the results of Table I, it is shown that, on average, the results of the different datasets are directly correlated to the accuracy of the network. As the standard deviation of the exposure increases, the mAP decreases. This can be stated, especially at dataset L4, which shows a low standard deviation and an overall good detection performance of 60.2.

### A. Depth information

By comparing the depth image results with those of the colour images, two major differences can be observed (Fig.4). Although the test images have the same lighting conditions as the training images, the depth image shows a mAP of just 51.5 (Table II). The colour images, on the other hand, have a mAP over 70.9. Due to the trained information, the network learns about the plant that under diffuse light conditions, like colours

TABLE I. EXPOSURE RATING

| Light condition | Time | Standard deviation |
|---|---|---|
| Diffus Light (L1) | 9 am | 63.0 |
| Strong Direct Light (L2) | 10 am | 71.0 |
| Strong Direct Light (L3) | 1:30 pm | 72.2 |
| Weak Direct Light (L4) | 3:30 pm | 66.9 |
| Strong Direct Light (L5) | 6:30 pm | 73.5 |

or structures, the colour images have a higher segmentation accuracy compared to the depth images.

The second difference in the depth image results is their constancy. Under the same exposure as the training images, the colour images average a 28.8% higher segmentation accuracy. Under an intense and direct light influence (L2, L3, and L5), the depth image has an average 21.7% higher mAP. The colour pictures show a variation in their precision depending on the exposure, as the plant's colour tones vary with the different exposition. Fig.4 R1 shows the different exposure situations and the resulting colour and shadow variations. With a standard deviation of 1.85, the mAP values of the depth image are very consistent in comparison to the colour images with a standard deviation of 11.61. The constant mAP values show the independence of the depth image created by SfM from changing lighting conditions. This relative independence of the exposure leads to the highest mAP value when the most strong shadows are cast (L5). Due to depth information, the cabbage can be segmented reliably at a constant high level regardless of the exposure, the colour characteristics of the plant or the leaf structures.

### B. Influence of different colour spaces

In general, it is shown that with an increasing standard deviation of the L-value, the accuracy of the instance segmentation decreases independent of the colour space. The analysis of the mAP shows that the CIELAB colour space (55.9) has a higher mAP than the RGB (51.5) and HSV (51.1) colour spaces (Table II) on average. The difference is most significant at L4 with a weak shadow cast. Here the LAB colour space has a 14.2 % higher segmentation accuracy than the RGB colour space and an 18.6 % higher segmentation accuracy than the HSV colour space. Since the HSV colour space and the CIELAB colour space have a channel describing light intensity, the difference in segmentation accuracy can be explained by the division of colour information.

TABLE II. MAP COMPARSION DEPTH LAYER INFLUENCE WITH HOMOGENE DATASET

| Light condition | Depth (mAP) | RGB (mAP) | RGB-D (mAP) | HSV (mAP) | HSV-D (mAP) | LAB (mAP) | LAB-D (mAP) | Overall mAP |
|---|---|---|---|---|---|---|---|---|
| Diffus Light (L1) | 51.5 | 70.9 | 71.7 | 71.9 | 73.6 | 74.3 | **75.0** | **72.5** |
| Strong Direct Light (L2) | 52.4 | 45.9 | 49.4 | 48.2 | 55.7 | 51.2 | **58.6** | 47.7 |
| Strong Direct Light (L3) | 52.1 | 46.0 | 47.1 | 38.0 | 53.4 | 44.9 | **54.9** | 43.7 |
| Weak Direct Light (L4) | 56.6 | 57.4 | 63.0 | 56.7 | 65.3 | 66.9 | **69.7** | 60.2 |
| Strong Direct Light (L5) | **54.3** | 37.3 | 48.3 | 40.6 | 48.8 | 42.0 | 50.2 | 42.5 |
| Average | 53.4 | 51.5 | 55.9 | 51.2 | 59.3 | 55.9 | **61.7** | |

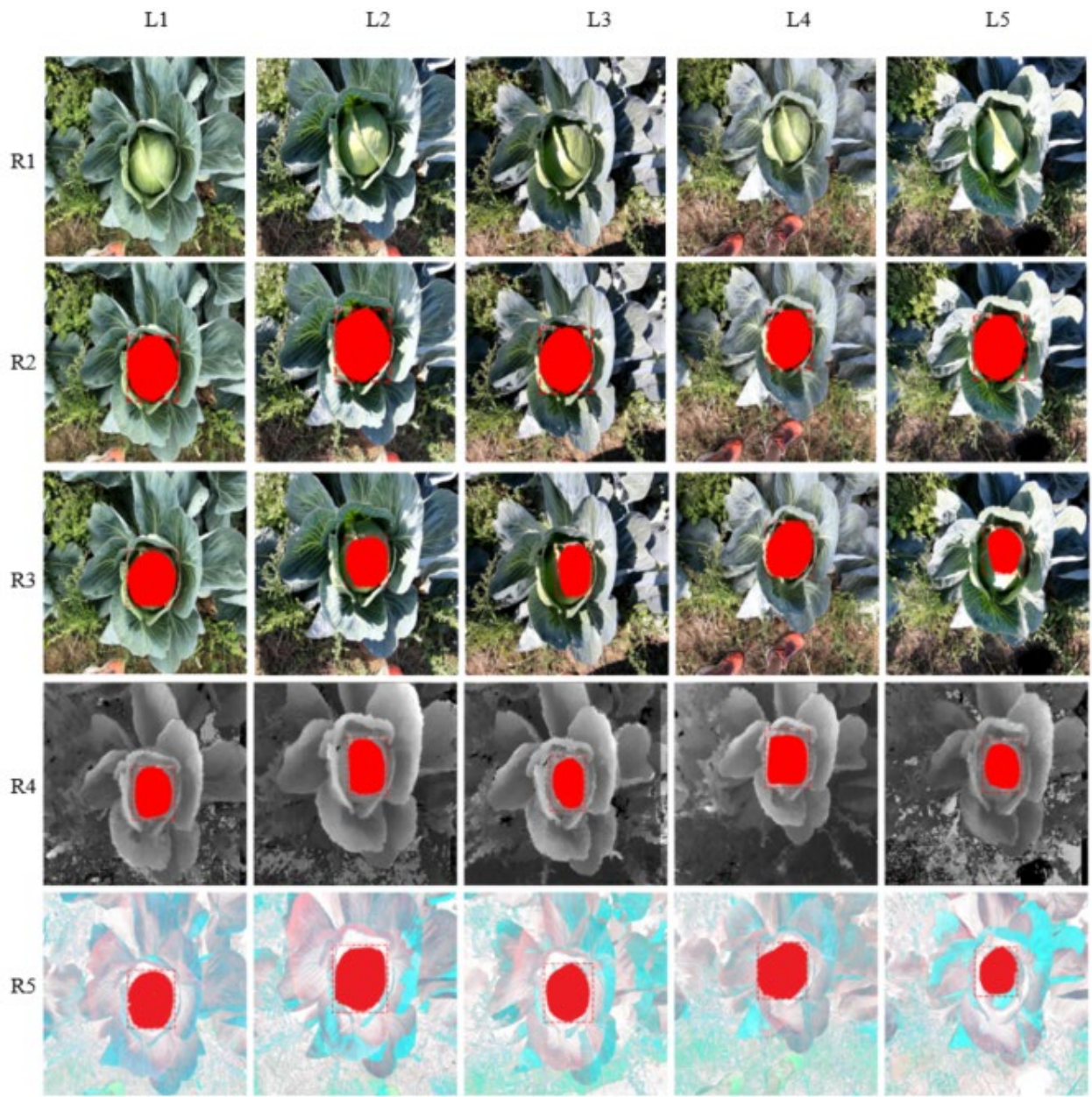

Fig.4: Segmentation results under different exposure situations (L1-L5): R1: RGB-Images, R2: Ground truth, R3: RGB-image segmentation results, R4: Depth-image segmentation results, R5: RGBD-image segmentation results.

### C. Results of colour and depth image

By adding a fourth image channel consisting of depth information to a colour image, it was investigated whether the exposure resistance of the depth image and the colour and structure information of the colour image can be combined. The results in Table II show that regardless of the colour channel, adding a fourth image channel with depth information improves instance segmentation accuracy and produces more robust results. Comparing the influence of depth information in different exposure scenarios shows that the segmentation accuracy for the RGB colour space increases by 10.2%, for the HSV colour space by 18%, and for the CIELAB colour space by 12.7% if an additional channel of depth is provided to the segmentation process. The most significant improvement through depth information was achieved in the HSV colour space with exposure scenario L3 with 28.8%. If the segmentation results (Fig.4) of the RGB-D image (R5) are compared with the segmentation results of the RGB (R3) and depth image (R4) under the exposure conditions of L3 and L5, it can be observed that the segmentation does not fully cover the cabbage in the RGB and depth image. The segmentation result of the RGB-D image shows a more precise detection of the cabbage in visual comparison with the ground truth images (R2). The CIELAB colour space with additional depth information (LAB-D) achieves the highest mAP values in four of the five exposure scenarios.

## V. Conclusion

This paper presents an approach to provide the most information-rich database for cabbage detection and segmentation. We have shown that it was possible to create a training dataset with a simple sensor setup that contains both; colour and depth information and thereby provides a more robust instance segmentation for varying environmental situations. The segmentation results of the depth images remained at a constant level under the five different exposure scenarios. The CIELAB colour space with an additional depth layer has proven to be the most information-rich, with an overall performance of 61.7 mAP.

## Acknowledgment

The DiWenkLa-project supported this research by funds of the Federal Ministry of Food and Agriculture (BMEL) based on a decision of the Parliament of the Federal Republic of Germany via the Federal Office for Agriculture and Food (BLE) under the innovation support program.